\documentclass{midl} 

\usepackage{mwe} 
\usepackage{rotating}
\usepackage{multirow}
\usepackage{booktabs}
\usepackage{tikz}
\usepackage{pgfplots}
\usepackage{algorithm}
\usepackage{algorithmic}
\usetikzlibrary{matrix}
\usetikzlibrary{positioning}
\usetikzlibrary{shapes.geometric}
\usetikzlibrary{arrows.meta,positioning,shapes,calc}
\jmlryear{2026}
\jmlrworkshop{Full Paper -- MIDL 2026}
\jmlrvolume{315}
\jmlrpages{3542--3560}
\editors{Accepted for publication at MIDL 2026}

\title[Task-Conditioned 3D U-Nets via Hypernetworks]{Task-Conditioned 3D U-Nets via Hypernetworks for Data-Scarce Medical Segmentation}

\midlauthor{\Name{Luca Hagen\nametag{$^{1}$}} \orcid{0009-0005-0990-6807} \Email{luca.hagen@fau.de}\\
\Name{Johanna P. Müller\nametag{$^{1}$}} \orcid{0000-0001-8636-7986} \Email{johanna.paula.mueller@fau.de}\\
\Name{Moritz Gmeiner\nametag{$^{1}$}} \Email{moritz.gmeiner@fau.de}\\
\Name{Bernhard Kainz\nametag{$^{1,2}$}} \orcid{0000-0002-7813-5023} \Email{bernhard.kainz@fau.de}\\
\addr $^{1}$ Friedrich-Alexander University Erlangen-Nuremberg, GER \\
\addr $^{2}$ Imperial College London, UK\\
}

\begin{document}

\maketitle

\begin{abstract}
Training 3D segmentation models typically requires extensive expert annotation, which is costly and often unavailable for rare or low-prevalence pathologies. We propose a hypernetwork-based framework that amortises the prediction of parameters for compact 3D U-Nets, enabling task-specific specialisation from as little as a single annotated volume. By learning shared anatomical structure, such as coarse shape, scale, and spatial organisation, across organs and imaging modalities, the hypernetwork generates task-conditioned network parameters, allowing controlled adaptation to previously unseen but anatomically related targets without full retraining.
We evaluate the proposed approach on the CT TotalSegmentator and Medical Segmentation Decathlon benchmarks. The method achieves strong one-shot performance for anatomically homogeneous structures (\emph{e.g.}, liver, spleen, atrium) and demonstrates stable few-shot adaptation for more heterogeneous or low-contrast targets (\emph{e.g.}, tumours, prostate). In regimes with two to four annotated volumes, hypernetwork-generated U-Nets consistently outperform pretrained baselines and substantially reduce the performance gap to fully supervised models while using minimal annotation. These results indicate that weight prediction serves as an effective task-informed prior for data-scarce 3D medical image segmentation.

\end{abstract}


\par\vspace{0.75\baselineskip}
\noindent\textit{Published in the Proceedings of the 9th International
Conference on Medical Imaging with Deep Learning (MIDL 2026),
PMLR 315, pp.~3542--3560.}
\href{https://proceedings.mlr.press/v315/hagen26a.html}{Original publication on PMLR}
\par

\smallskip
\noindent\textbf{Code.}
\href{https://github.com/luca-hagen/HyperUNet}{Preliminary implementation}
(repository under development).
\par

\section{Introduction}
Automated segmentation is a core component of many radiological workflows, with neural networks enabling accurate delineation of organs, tumours, and other anatomical structures. Architectures such as the U-Net \cite{ronneberger2015u} and, more recently, Vision Transformer variants \cite{dosovitskiy2020image} remain widely used for 3D medical image segmentation. Despite their success, these models typically rely on large amounts of expert-annotated data, which limits their applicability for rare pathologies or uncommon anatomical targets.
The standard supervised learning pipeline requires extensive data collection, careful curation, and time-consuming expert annotation \cite{galbusera2024annotation}, followed by task-specific model training. While feasible for common and routinely imaged conditions, this paradigm breaks down when target cases are scarce. In such settings, the annotation burden represents a substantial entry barrier, despite the potential clinical value of automation.
One-shot and few-shot segmentation methods aim to address this challenge by conditioning a generic model on a small number of annotated reference examples. These approaches enable generalisation to unseen anatomical targets without retraining. Indeed, meta-learning-based segmentation has already been explored in medical imaging \cite{khadka2022meta,farshad2022metamedseg,leng2024self,alsaleh2024few,tirpude2025meta}. More broadly, few-shot and low-shot learning in medical imaging has become an active research area, as reviewed in recent surveys \cite{pachetti2024systematic,dissanayake2025few}. However, despite their flexibility, these methods exhibit two fundamental limitations.

First, their performance typically remains below that of models trained specifically for a given task, even when additional annotated examples become available. Their ability to generalise across tasks often comes at the cost of limited capacity for task-specific optimisation. Second, fine-tuning is non-trivial, as task conditioning is often realised through shared internal representations or logit modulation rather than explicit parameter adaptation. As a result, updating model weights for a single task can degrade performance on others, limiting their usefulness when more data is acquired.
Other lines of work attempt metric- or embedding-based segmentation under extreme scarcity, showing promising results even in one-shot scenarios \cite{9298830}. Approaches that rely on self-supervision and anomaly detection have also been proposed to mitigate the problem of limited foreground/background discrimination in few-shot settings \cite{hansen2022anomaly}. These efforts demonstrate the growing recognition of data scarcity in medical segmentation and the need for more robust, adaptable methods.
We address these limitations by formulating task conditioning as an explicit parameter-prediction problem. We propose a hypernetwork that amortises the generation of parameters for compact 3D U-Nets, conditioned on limited annotated data. Trained across a diverse set of segmentation tasks, the hypernetwork learns shared anatomical structure such as coarse shape, scale, and spatial organisation. By predicting weights rather than modulating logits or embeddings, the approach produces fully instantiated, task-specific U-Nets that are decoupled from the hypernetwork after generation.
This separation allows the generated U-Nets to be fine-tuned using standard optimisation as additional data becomes available, without interfering with the learned task prior or other tasks. As a result, the proposed method combines the data efficiency of few-shot learning with the robustness and interpretability of conventional segmentation models, enabling controlled adaptation to anatomically related targets under limited supervision.

\noindent\textbf{Contributions.} This work makes the following contributions:  
(1) A hypernetwork framework that generates task-specific 3D U-Net parameters from minimal annotated data for one- and few-shot segmentation of unseen targets.  
(2) Task conditioning via explicit weight prediction, producing fully instantiated U-Nets that can be fine-tuned independently of the hypernetwork.  
(3) Demonstration that the hypernetwork captures shared anatomical structure across organs and modalities, providing task-informed priors for data-efficient segmentation.  
(4) Extensive evaluation on CT TotalSegmentator and the Medical Segmentation Decathlon shows that \textsc{HyperUNet} achieves strong one- and few-shot performance across diverse anatomical targets with substantially reduced annotation requirements. 

Unlike SAM-style prompting and in-context learning methods that perform reference-conditioned inference at test time, \textsc{HyperUNet} synthesises a standalone, task-specific 3D U-Net, enabling efficient deployment and incremental fine-tuning without increasing inference-time memory or computation.

\section{Background}
Few-shot models have become a central strategy for medical image segmentation under data scarcity. Given only a few annotated reference volumes, these models adapt to new tasks by exploiting information extracted from the support set. Existing approaches can be broadly grouped into three paradigms.
Similarity- and prototype-based methods classify query voxels by matching pixel-wise embeddings to prototypes constructed from the support set \cite{9298830}. While effective for anatomically homogeneous structures, their performance depends heavily on the expressiveness of the embedding space. As a result, they struggle with complex 3D anatomy, high intra-class variability, and the inherently difficult foreground-background separation in low-shot regimes \cite{9298830, hansen2022anomaly}.
Attention-based models use cross-attention to align query features to support features \cite{galbusera2024annotation, hu2025medverse}. This enables strong appearance transfer but tightly couples the inference procedure to the support examples. Since these models do not instantiate standalone segmentation models, they cannot be easily fine-tuned or deployed independently of the conditioning samples.
Parameter-based meta-learning focuses on learning how to adapt a segmentation model from few reference examples. MAML~\cite{finn2017model} and its variants \cite{khadka2022meta, leng2024self, alsaleh2024few} learn shared initialisations that are rapidly fine-tuned to new tasks. Volumetric extensions show generalisation across heterogeneous targets \cite{farshad2022metamedseg, tirpude2025meta}. However, because task adaptation is restricted to a few gradient steps, the resulting models remain bound to a local neighbourhood of the meta-initialisation, ultimately limiting their flexibility.
First introduced in \cite{ha2016hypernetworks}, hypernetworks generate the parameters of another model directly in a forward pass. They have been used for fast task-specific updates \cite{przewikezlikowski2024hypermaml}, full model parameterisations \cite{zhmoginov2022hypertransformer}, and sample-conditioned filter generation \cite{nirkin2021hyperseg}. Unlike MAML-based methods, which rely on solutions near a shared initialisation, hypernetworks can represent a broader family of task-specific models.
This distinction matters in few-shot 3D segmentation. MAML requires multiple fine-tuning steps, substantial per-task supervision, and operates on closely related tasks and organs. Our approach instead generates a complete task-specific U-Net in a single forward pass and applies only one light update on the generated model, leaving the hypernetwork unchanged. This enables stable, modular adaptation and effective generalisation to genuinely unseen anatomical targets from as little as one annotated volume. Motivated by these advantages, we adopt a hypernetwork to predict autonomous, fine-tunable segmentation models.
This \emph{generate-once} paradigm also differentiates our method from prominent in-context learning segmentation frameworks, such as \emph{UniverSeg}~\cite{butoi2023universeg} and \emph{Iris}~\cite{gao2025show}, as well as promptable segmentation approaches such as \emph{MedSAM}~\cite{cheng2023sam,wang2025sam}.
Similar to attention-based retrieval mechanisms, these methods maintain a tight coupling to a reference set at inference time: predictions are obtained by explicitly conditioning on, or comparing against, the provided context examples.
As a consequence, incorporating additional supervision is typically non-trivial and often requires re-running the full inference procedure with an expanded context.

In contrast, our approach compiles decision rules from a reference sample \emph{once} in the form of a full U-Net decoder, yielding a standalone segmentation model that can be deployed independently of the original reference set.
Crucially, when further annotations become available, the resulting model can be adapted using standard backpropagation, without modifying the underlying hypernetwork pipeline.

\section{Method}
To tackle data-scarce 3D medical image segmentation, we propose a hypernetwork framework that generates task-specific parameters for compact 3D U-Nets. Instead of feature- or logit-based conditioning, our method treats task adaptation as explicit weight prediction, producing fully instantiated U-Nets that can be fine-tuned independently. By training across diverse segmentation tasks, the hypernetwork captures shared anatomical structures, such as shape, scale, and spatial organisation, enabling adaptation to unseen but anatomically related targets with minimal annotated data. We first formalise the problem and then detail the architecture, training strategy, and task-specific parameter generation.

\paragraph{Problem Formulation.}

We focus on binary 3D medical image segmentation tasks. Each task $\mathcal{T}$ is defined by a dataset $D = \{(x_i, y_i)\}_{i=1}^N$, where $x_i \in \mathbb{R}^{H \times W \times D}$ is a 3D image volume and $y_i \in \{0,1\}^{H \times W \times D}$ is its corresponding binary mask.
For multi-class datasets with labels $y_i \in \{0,\dots,K\}^{H \times W \times D}$, we decompose the problem into $K$ binary segmentation tasks $\{\mathcal{T}_k\}_{k=1}^K$, one per foreground class. When volumes have multiple channels (e.g., different MRI sequences), each channel-class combination is treated as a separate binary task.
Given a small support set of one or a few reference volume-mask pairs for a task $\mathcal{T}$, our goal is to predict the weights of a compact 3D U-Net that can be deployed as an autonomous segmentation model for that task. By formulating task adaptation as explicit parameter prediction, rather than feature- or logit-based conditioning, we aim to produce models that are both task-specific and fine-tunable without affecting other tasks.

\paragraph{Target Architecture.}

Our target network is a compact 3D U-Net composed of four encoder and four decoder stages, using a residual double-convolution block (ResDoubleConv) as the primary building unit. Each block contains two $3{\times}3{\times}3$ convolutions with group normalisation and LeakyReLU activations, alongside a $1{\times}1{\times}1$ residual branch for stable gradient flow.
Downsampling in the encoder is implemented via strided $3{\times}3{\times}3$ convolutions, while upsampling in the decoder uses transposed convolutions. To reduce the number of parameters, skip connections between the encoder and decoder stages are additive rather than concatenative. A final $1{\times}1{\times}1$ convolution maps decoder features to segmentation logits.
Across tasks, the encoder, normalisation layers, and upsampling layers are shared and trained jointly. Task-specific adaptation is achieved by a hypernetwork that generates all convolutional weights in the decoder ResDoubleConv blocks and the final output head. This design produces compact, fully instantiated U-Nets with approximately 5.7 million trainable parameters, combining efficiency, interpretability, and the ability to fine-tune per task without affecting other tasks.

\paragraph{Hypernetwork-based U-Net Generation.}

\begin{figure}[h!]
    \centering
    \includegraphics[width=\textwidth]{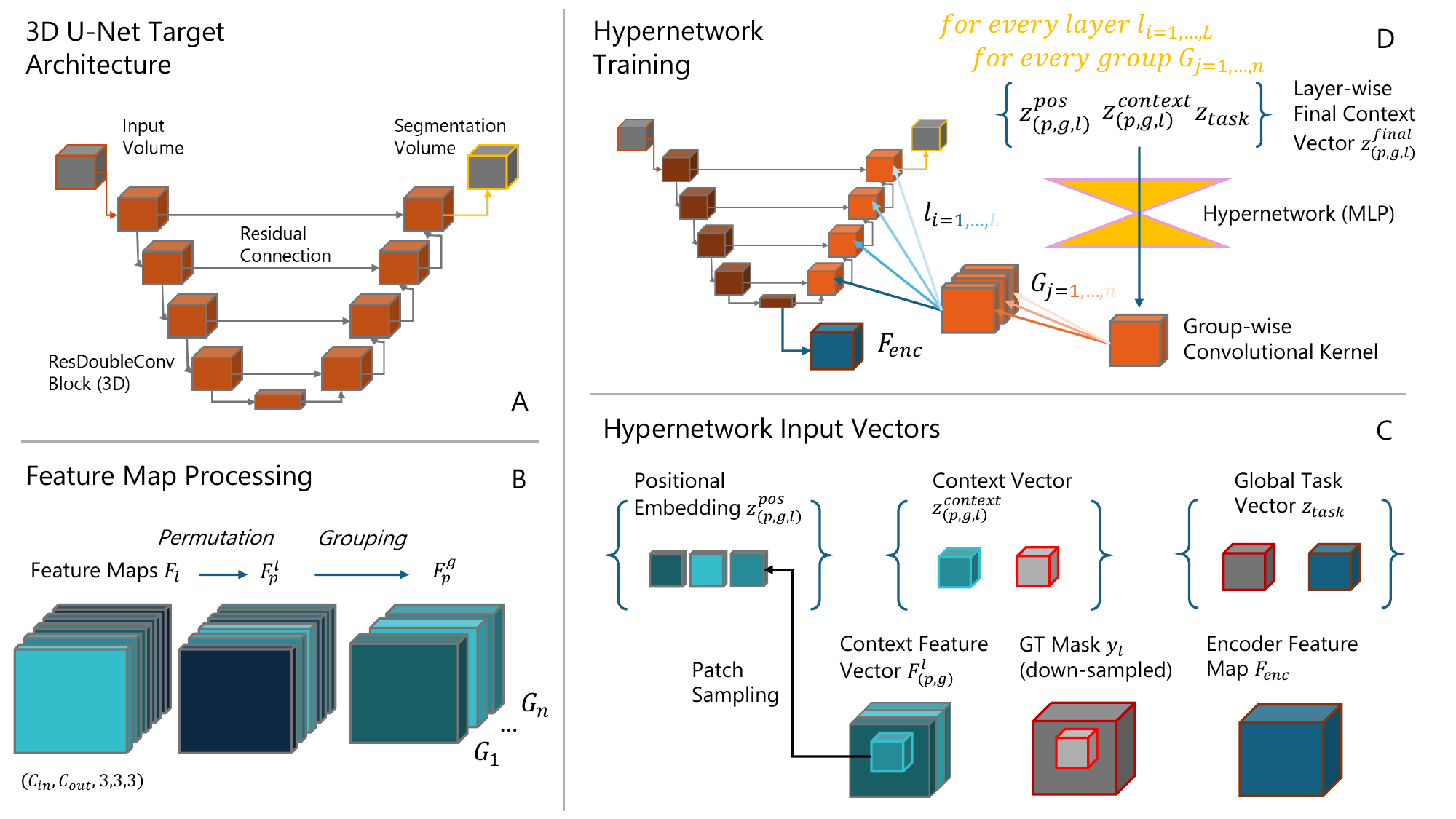}
    \caption[Context construction]{\textbf{Hypernetwork-driven U-Net Parametrisation}. (A) Target 3D U-Net for segmentation. (B) Feature map processing for generation of Context Feature Vectors for each layer. (C) Generation of all necessary vectors and positional embeddings for patch sampling. (D) Hypernetwork Training with shared decoder weights and predicted encoder weights from the hypernetwork, given the final context vector for each decoder layer.}
    \label{fig:context_construction}
\end{figure}

Given a reference volume-mask pair $(x, y)$ for a task $\mathcal{T}$, the hypernetwork generates all decoder weights of the corresponding target U-Net in a single forward pass. This allows the resulting U-Net to be fully instantiated and fine-tuned independently of the hypernetwork.
We first encode the input volume $x$ using the shared encoder $E$ to obtain a high-level feature map $F_{\text{enc}} = E(x) \in \mathbb{R}^{C \times H' \times W' \times D'}$. To summarise the task, we extract a global descriptor from the reference annotation by applying mask average pooling over the foreground region:
\begin{equation}
z_{\text{task}} = \mathrm{AvgPool}\big(\mathrm{Upsample}(F_{\text{enc}}) \odot y\big) \in \mathbb{R}^{1 \times C},
\end{equation}
where $\odot$ denotes elementwise multiplication, and $\mathrm{Upsample}(\cdot)$ resamples the feature map back to the original resolution. This descriptor $z_{\text{task}}$ encodes the spatial and semantic information of the task and is subsequently used by the hypernetwork to generate decoder weights.
\subsection{Context Vector Construction for Decoder Convolutions.}
In our setup, the weights of each convolutional layer are generated by a forward pass of a reference pair $(x,y)$ and on demand, i.e., exactly at the moment when the respective layer is required during the forward pass. Consequently, the convolutional layers of the decoder are instantiated \textbf{sequentially}.

First, we compute the encoder feature map $F_{enc}$, it can be directly forwarded into the decoder. At this stage, the convolutional layers of the decoder are all uninitialized, while non-parametric components such as upsampling and normalisation layers can be applied as usual.
Following the standard U-Net decoder structure, the first convolutional decoder layer is encountered after the initial upsampling operation and the incorporation of the corresponding encoder skip connection. We denote the resulting feature map, i.e., the input that would normally be passed through the convolutional layer, as $F^l$. In our notation later, feature maps $F_l$ will be indexed by their respective convolutional decoder layer $l \in L$.
In the following, we describe the context generation process for a decoder convolutional layer.
In our setup, each generated convolutional layer have matching input/output channel size (in-channels = out-channels). Let $C_{l}$ denote the channel size of our current convolutional layer, then we have to generate a weight tensor of shape $(C_{l}, C_{l}, 3, 3, 3)$. Instead of generating $27 \times C_{l}^2$ parameters directly, we generate them in chunks of size $27 \times C_{base}^2$, where $C_{base}$ is a hyperparameter such that $C_{l} \mod C_{base} = 0$ (in our setup $C_{base}$ is chosen to be the same as the number of our U-Net base-channels; 24). Consequently, we have to generate $G = C_{l} / C_{base}$ different chunks of weights, where each chunk corresponds to one context vector and is obtained by passing the respective context vector through the hypernetwork.

\paragraph{Permutations.} To construct the context vectors, we first create $G$ copies of $F_{l}$ and permute each of them individually over the channel dimension. 

\paragraph{Channel partitioning.} Each of these permuted feature maps $\{F^l_{p}\}_{p=1}^G$, are then further partitioned into $G$ non-overlapping groups along the channel dimension. We denote the resulting groups as $F^{g}_{p} \in \mathbb{R}^{C_{\mathrm{base}}\times h\times w\times d}$, with $p \in \{1, ..., G\}$ and $g \in \{1, ..., G\}$. Note that due to the permutation step, each of the $C_{base}^2$ different $F^l_{(p,g)}$'s is a unique combination of channels from the feature map $F_{in}$.

\paragraph{Mask Downsampling.} As a next step, we downsample the binary support mask $Y$ to match the spatial dimension of our $F^l_{(p,g)}$'s with $y_{down} = \mathrm{Downsample}(y) \in \{0,1\}^{h \times w \times d}$.

\paragraph{Patch Sampling.} For each $F^l_{(p,g)}$ we now sample $N$ local patches of size $3 \times 3 \times 3$ 
\begin{equation}
p^{(i)}_n \in \mathbb{R}^{C_{\mathrm{base}}\times 3 \times 3 \times 3}, 
\qquad n=1,\dots,N,
\end{equation}
from $F^l_{(p,g)}$. We sample such that $N/2$ of the resulting patches are positive and $N/2$ are negative. Additionally, our sampling is boundary-aware: For both, negative and positive patches, we ensure that at least a fixed fraction $b$ ($0.7$ in our experiments) of them is a boundary patch, meaning that when looking at the respective mask-patch from $y_{down}$, both foreground and background are present. Positive and negative patches are labelled with respect to their centre voxel in $y_{down}$, meaning a negative patch has a 0 in its centre voxel when looking into the respective $3 \times 3 \times 3$ patch in $y_{down}$.
The rationale behind using local patches is to show our hypernetwork multiple small local examples of foreground and background regions, while laying special emphasis on boundary regions, which, as we argue, hold the most information for distinction. In comparison to using voxels, patches have the advantage that they not only carry information about features but also the feature-gradient.
We stack the resulting patches in a fixed order [positive → positive-boundary → negative → negative-boundary]. Flattening and concatenating the patches yields the context vector $z^{\mathrm{context}}_{(p,g,l)} \in \mathbb{R}^{C_{\mathrm{base}}\cdot 27 \cdot N}$ of $F^l_{(p,g)}$.

\paragraph{Task-aware positional embedding}. While each of the $z^{\mathrm{context}}_{(p,g,l)}$'s now carries specific information how to seperate foreground from background for its group $F_p^g$, it does not hold any information about the position in which its resulting weights will be employed in, in the decoder architecture. To enrich the context vector $z$ with such positional information, we assign a learnable positional embedding $z^{pos}_{(p,g,l)}$ to each group $F^l_{(p,g)}$ from decoder layer $l$. As we argue, that each group $F^{g}_{p}$ can have different roles for different tasks, we enrich these $z^{pos}_{(p,g,l)}$ with our global task vector $z_{task}$ by merging them through a small two-layer MLP:
\begin{equation}
    z^*_{(p, g, l)} = \mathrm{MLP(Concat[z_{task}, z^{pos}_{(p,g,l)}]}.
\end{equation}

\paragraph{Final context vector.} Finally, we can give the context-, position- and task-aware input vector for our hypernetwork as:
\begin{equation}
    z^{final}_{(p, g, l)} = \mathrm{Concat}[z_{p,g}^{context}, z^*_{(p, g, l)}]
\end{equation},
which are then passed into the hypernetwork to generate the weights for the layer l.

\paragraph{Hypernetwork.}
The hypernetwork $H$ is implemented as a three-layer MLP with one hidden layer of dimension 2048 and GELU activations. It maps each final context vector $z^{final}_{(p, g, l)}$ to a base convolutional filter of shape $(C_{\mathrm{base}}, C_{\mathrm{base}}, 3, 3, 3)$. Layer normalisation is applied after the first two linear layers to stabilise training and improve convergence.
Once all base convolutional filters for layer $l$ are generated, they are stacked along the input and output channel dimensions to produce a full weight matrix $W_l$ for the convolutional kernel of shape $(C_{\mathrm{out}}, C_{\mathrm{in}}, 3, 3, 3)$, with $C_{\mathrm{in}} = C_{\mathrm{out}} = G \cdot C_{\mathrm{base}}$.
As soon as $W_l$ is obtained, we can pass the current feature map $F_l$ through the convolution parameterised by $W_l$. The obtained feature map $F_{out}$ is further passed through the decoder until the next uninitialised layer is reached. 
The weights and the bias of the final $1{\times}1{\times}1$ segmentation head are generated by a separate two-layer MLP conditioned on the global task descriptor $z_{\mathrm{task}}$. This design produces a fully instantiated decoder that is both task-specific and ready for independent fine-tuning.
Once a forward pass through the entire decoder is complete, all layers are initialised, and then we can treat the model as a standard 3D U-Net.

\paragraph{Reference-based weight update.}
Once the decoder weights are generated, they are combined with the shared encoder to obtain a complete task-specific U-Net. To further leverage the information from the reference pair $(x, y)$, we perform a single gradient update on the U-Net parameters using a small learning rate, while keeping the hypernetwork fixed. The resulting model is a fully instantiated, task-specific U-Net that can be used directly for inference or as a task-informed initialisation for further fine-tuning when additional labelled data becomes available. This separation ensures that the hypernetwork retains its ability to generalise across tasks, while the generated U-Net can specialise to the current task.

\begin{algorithm}[H]
\caption{Episodic Training for Hypernetwork-predicted U-Nets}
\label{alg:episodic_training}
\begin{algorithmic}[1]
\STATE Initialize episode counter $episode = 1$
\WHILE{$episode \leq N_\text{episodes}$}
    \STATE Sample a mini-batch $\mathcal{B}$ of tasks
    \FOR{each task $\mathcal{T}$ in $\mathcal{B}$}
        \STATE Randomly sample a support pair $(x_s, y_s)$ for task $\mathcal{T}$
        \STATE Generate task-specific U-Net: $F_\mathcal{T} = \text{Hypernetwork}(x_s, y_s)$
        \STATE Sample an independent query pair $(x_q, y_q)$ for the same task
        \STATE Predict outputs: $\hat{y}_s = F_\mathcal{T}(x_s)$, $\hat{y}_q = F_\mathcal{T}(x_q)$
        \STATE Compute combined loss:
        \STATE \hspace{\algorithmicindent} $\mathcal{L} = \beta \, L_{\text{seg}}(\hat{y}_s, y_s) + (1-\beta) \, L_{\text{seg}}(\hat{y}_q, y_q)$
    \ENDFOR
    \STATE Update parameters of $E$, $H$, and shared decoder layers using AdamW with exponentially decaying learning rate
    \STATE $episode = episode + 1$
\ENDWHILE
\STATE Apply early stopping based on validation performance on unseen tasks
\end{algorithmic}
\end{algorithm}

\paragraph{Episodic Training Objective}

We train the encoder $E$, hypernetwork $H$, and shared decoder components end-to-end in an episodic meta-learning setup. Each episode contains multiple tasks with support-query pairs. The training procedure is summarised in Algorithm~\ref{alg:episodic_training}.

\section{Experiments}
We assess our framework on tasks entirely unseen during training by generating a \textsc{HyperUNet} for each new target. Performance is compared against three baselines sharing the same U-Net architecture: (i) a conventionally trained, fully supervised model for the given target task (Conv U-Net; upper bound), (ii) a model using the shared encoder, upsampling, and normalisation layers but randomised decoder weights (Rand U-Net), and (iii) a model trained jointly on all tasks including target tasks (All-classes U-Net) across both datasets, covering over 130 anatomical targets across CT and multiple MRI sequences. The All-classes U-Net baseline serves to assess whether a single multi-task model with comparable capacity can absorb knowledge across tasks under partial labels, and to contrast this with our task-conditioned decoder generation. In addition, we report three ablation studies for our method, detailed in Appendix~\ref{appendix}.

\paragraph{Datasets.}
We train and evaluate on two public benchmarks selected for anatomical diversity and suitability for episodic meta-learning. CT TotalSegmentator~\cite{wasserthal2023totalsegmentator} provides voxel-wise annotations for over 100 thoraco-abdominal structures and is used to define a large set of binary training tasks spanning organs, vessels, bones, and soft tissue. The Medical Segmentation Decathlon (MSD~\cite{antonelli2022medical} consists of 10 heterogeneous CT and MRI datasets covering diverse anatomies and pathologies. In MSD, all evaluation targets are defined as hold-out tasks that are never used during hypernetwork training and serve exclusively for out-of-distribution evaluation across both CT and MRI domains.

\paragraph{Preprocessing and Training.}
All volumes are resampled to isotropic $1.5\,\mathrm{mm}$ spacing and reoriented to the RAS coordinate system. CT intensities are clipped to $[-900,900]$ and Z-score normalised, while MRI volumes are normalised by mapping the 1st to 99th intensity percentiles and clipping outliers. A single $128^3$ voxel patch covering the target structure is extracted per volume and centred when necessary. On-the-fly spatial (flips, $90^\circ$ rotations, affine transforms) and intensity augmentations (Gaussian noise, smoothing, contrast adjustments) are applied. Models are trained for 20,000 episodes using a combined Dice and binary cross-entropy loss with $\beta=0.2$, optimised with AdamW and an exponentially decaying learning rate. Early stopping is based on validation performance on unseen tasks. Training is performed on four NVIDIA A100 GPUs for approximately 12 hours.

\paragraph{CT Domain.}
First, we investigate the capabilities of our approach on unseen CT tasks. As our model was trained exclusively in the CT domain, this experiment examines the models' abilities to generalise to adapt to unseen organs / new tasks.
\begin{table*}[t]
\centering
\caption[CT Domain Results]{\textbf{Organ- and Task-Shift results}. Dice (DSC) and Normalised Surface Dice (NSD) for Random U-Net, Conventional U-Net, and \textbf{\textit{1-shot}} \textsc{HyperUNet} (Ours) across all CT tasks. *Models trained on target tasks. \textbf{Best} and \underline{second-best} per task are highlighted.}
\resizebox{\textwidth}{!}{%
\begin{tabular}{l cc cc cc}
 
& \multicolumn{2}{c}{\textbf{Liver}} 
& \multicolumn{2}{c}{\textbf{Spleen}} 
& \multicolumn{2}{c}{\textbf{Hepatic Vessels}} \\
\cmidrule(lr){2-3} \cmidrule(lr){4-5} \cmidrule(lr){6-7}
\textbf{Network}& DSC & NSD & DSC & NSD & DSC & NSD \\
\cmidrule(lr){2-2} \cmidrule(lr){3-3}\cmidrule(lr){4-4}\cmidrule(lr){5-5}\cmidrule(lr){6-6}\cmidrule(lr){7-7}
Rand. U-Net 
& 0.176$\pm$0.050 & 0.110$\pm$0.025 
& 0.091$\pm$0.052 & 0.045$\pm$0.017 
& 0.019$\pm$0.010 & 0.032$\pm$0.012 \\
All-classes U-Net* 
& \textbf{0.826$\pm$0.073} & \textbf{0.668$\pm$0.106} 
& \textbf{0.580$\pm$0.167} & \textbf{0.509$\pm$0.123} 
& \underline{0.050$\pm$0.023} & \underline{0.038$\pm$0.114} \\
\textsc{HyperUNet} \emph{(Ours)} 
& \underline{0.772$\pm$0.063} & \underline{0.490$\pm$0.080}
& \underline{0.501$\pm$0.169} & \underline{0.316$\pm$0.115}
& \textbf{0.307$\pm$0.120} & \textbf{0.364$\pm$0.129} \\
\emph{Conv. U-Net*} 
& \emph{0.951$\pm$0.018} & \emph{0.792$\pm$0.069} 
& \emph{0.911$\pm$0.032} & \emph{0.708$\pm$0.108} 
& \emph{0.607$\pm$0.104} & \emph{0.731$\pm$0.103} \\
\end{tabular}%
}
\resizebox{\textwidth}{!}{%
\begin{tabular}{l cc cc cc}
& \multicolumn{2}{c}{\textbf{Pancreas}} 
& \multicolumn{2}{c}{\textbf{Lung Tumor}} 
& \multicolumn{2}{c}{\textbf{Colon Tumor}} \\
\cmidrule(lr){2-3} \cmidrule(lr){4-5} \cmidrule(lr){6-7}
\textbf{Network} & DSC & NSD & DSC & NSD & DSC & NSD \\
\cmidrule(lr){2-2} \cmidrule(lr){3-3}\cmidrule(lr){4-4}\cmidrule(lr){5-5}\cmidrule(lr){6-6}\cmidrule(lr){7-7}
Rand. U-Net 
& 0.016$\pm$0.007 & 0.017$\pm$0.006 
& 0.006$\pm$0.014 & 0.005$\pm$0.006 
& 0.012$\pm$0.017 & 0.010$\pm$0.010 \\
All-classes U-Net* 
& \textbf{0.282$\pm$0.171} & \textbf{0.174$\pm$0.132} 
& \textbf{0.120$\pm$0.181} & \textbf{0.079$\pm$0.044} 
& \textbf{0.070$\pm$0.115} & \textbf{0.046$\pm$0.098} \\
\textsc{HyperUNet} \emph{(Ours)} 
& \underline{0.153$\pm$0.138} & \underline{0.102$\pm$0.078}
& \underline{0.025$\pm$0.067} & \underline{0.019$\pm$0.058}
& \underline{0.050$\pm$0.126} & \underline{0.046$\pm$0.109} \\
\emph{Conv. U-Net*} 
& \emph{0.776$\pm$0.076} & \emph{0.632$\pm$0.118} 
& \emph{0.439$\pm$0.268} & \emph{0.340$\pm$0.307} 
& \emph{0.389$\pm$0.287} & \emph{0.277$\pm$0.171} \\
\end{tabular}%
}
\label{tab:results_ct_compact}
\end{table*}
We evaluate our hypernetwork on several tasks from the MSD dataset and report both the Dice Score (DSC) and the normalised surface distance (NSD) with a tolerance of $\tau = 2\,\mathrm{mm}$. As shown in Fig.~\ref{tab:results_ct_compact}, a consistent trend emerges: \textsc{HyperUNet}s substantially outperform the randomised baseline across all tasks, yet remain well below the performance of a fully supervised U-Net. Notably, \textsc{HyperUNet}s outperform the jointly trained All-classes U-Net on the hepatic vessels task (DSC 0.31 vs.\ 0.05), highlighting limitations of a single multi-task model under partial-label supervision. In this setting, rare and fine-grained structures such as hepatic vessels receive weak and infrequent supervision and are affected by gradient interference from competing tasks, leading to unstable or degraded performance. The strongest results are observed for larger and anatomically homogeneous structures such as the liver and spleen, indicating that the hypernetwork captures coarse shape and spatial priors effectively. For more complex or fine-grained structures (e.g., pancreas, hepatic vessels), performance drops relative to fully supervised training but still clearly exceeds a random baseline. For highly heterogeneous targets such as lung and colon tumours, both \textsc{HyperUNet}s and the All-classes U-Net show limited performance gains, reflecting the difficulty of learning robust tumour representations under partial labels and extreme inter-case variability when only a single support example is available.
\begin{figure}[t]
\centering
\begin{minipage}[t]{0.42\linewidth}
  \centering
  \includegraphics[width=\linewidth]{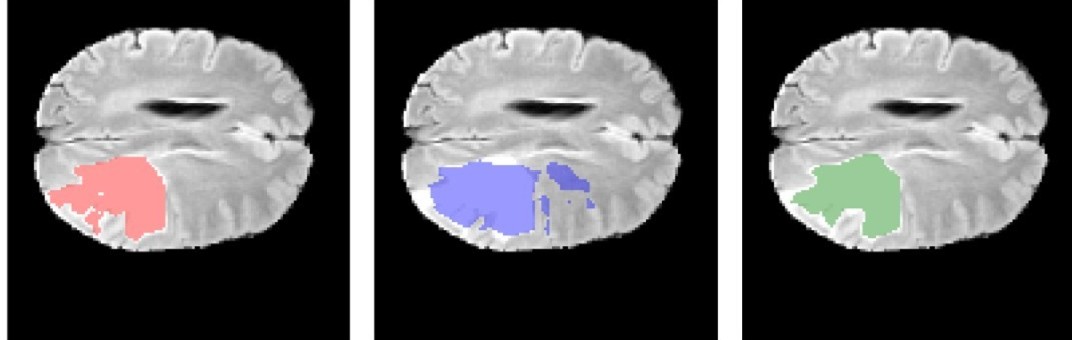}
  {\footnotesize (a) Brain Edema}
\end{minipage}
\hfill
\begin{minipage}[t]{0.42\linewidth}
  \centering
  \includegraphics[width=\linewidth]{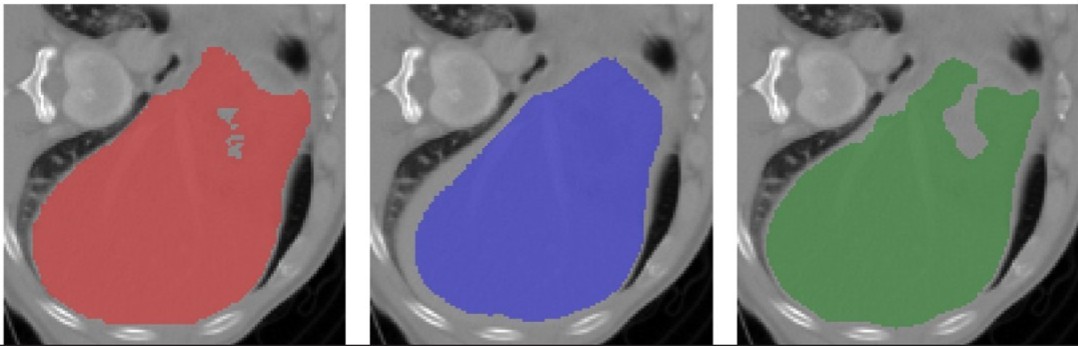}
  {\footnotesize (b) Liver}
\end{minipage}
\vspace{0.5em}
\begin{minipage}[t]{0.42\linewidth}
  \centering
  \includegraphics[width=\linewidth]{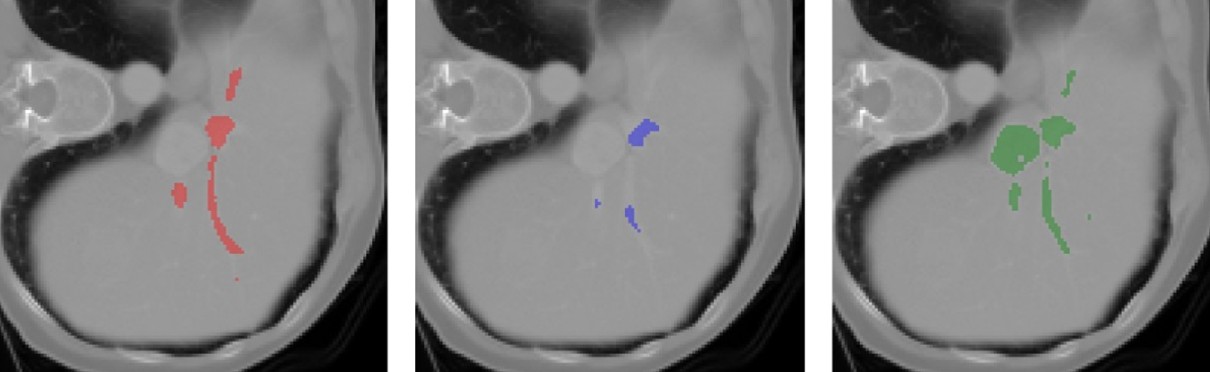}
  {\footnotesize (c) Hepatic Vessel}
\end{minipage}
\hfill
\begin{minipage}[t]{0.42\linewidth}
  \centering
  \includegraphics[width=\linewidth]{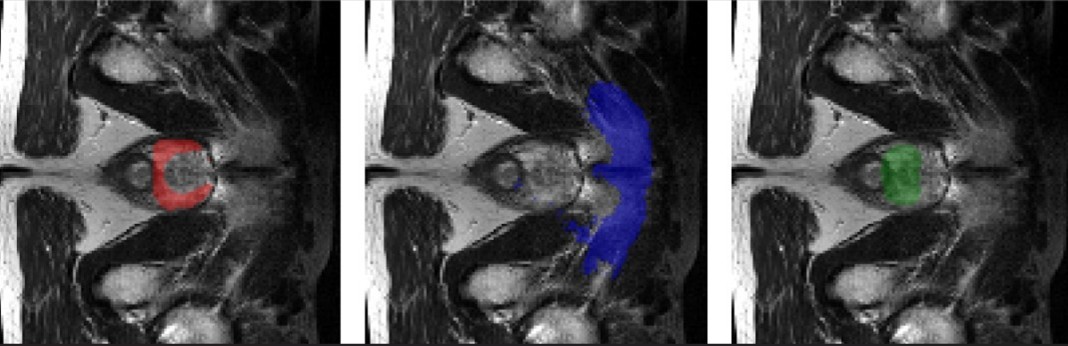}
  {\footnotesize (d) Prostate}
\end{minipage}
\caption{\textbf{Qualitative evaluation on four target-tasks.} Ground truth (red), and predictions of \textsc{HyperUNet} (blue) and \emph{Task-specific Conv. U-Net} (green).}
\label{fig:qualitative_examples}
\end{figure}

\paragraph{MRI Domain.}
In this experiment, in addition to a organ-/task-shift, we introduce a modality shift. Our hypernetwork approach was trained exclusively on CT tasks, but is now evaluated on unseen organs in the MRI domain, presenting a severe challenge.
\begin{table*}[t]
\centering
\caption[MRI Domain Results]{\textbf{Domain-, Organ-, and Task-Shift results.} Dice (DSC) and Normalised Surface Dice (NSD) for Random U-Net, Conventional U-Net, and \textbf{\textit{1-shot}} \textsc{HyperUNet} (Ours) across all MRI tasks. *Models trained on target tasks. \textbf{Best} and \underline{second-best} per task highlighted.}
\resizebox{0.8\textwidth}{!}{%
\begin{tabular}{l cc cc}
\textbf{Network} 
& \multicolumn{2}{c}{\textbf{Brain Edema (FLAIR)}} 
& \multicolumn{2}{c}{\textbf{Hippocampus (T1w)}} \\
\cmidrule(lr){2-3} \cmidrule(lr){4-5}
& DSC & NSD & DSC & NSD \\
\cmidrule(lr){2-3} \cmidrule(lr){4-5}
Rand. U-Net 
& 0.002$\pm$0.002 & 0.028$\pm$0.016 
& 0.002$\pm$0.001 & 0.001$\pm$0.001 \\
All-classes U-Net* 
& \textbf{0.508$\pm$0.195} & \textbf{0.436$\pm$0.167} 
& \underline{0.008$\pm$0.001} & \underline{0.001$\pm$0.002} \\
\textsc{HyperUNet} \emph{(Ours)} 
& \underline{0.326$\pm$0.208} & \underline{0.217$\pm$0.150} 
& \textbf{0.057$\pm$0.021} & \textbf{0.075$\pm$0.016} \\
\emph{Conv. U-Net*} 
& \emph{0.617$\pm$0.159} & \emph{0.584$\pm$0.109} 
& \emph{0.856$\pm$0.030} & \emph{0.970$\pm$0.034} \\
\end{tabular}%
}
\resizebox{0.8\textwidth}{!}{%
\begin{tabular}{l cc cc}
\textbf{Network} 
& \multicolumn{2}{c}{\textbf{Heart (bSSFP)}} 
& \multicolumn{2}{c}{\textbf{Prostate (T2)}} \\
\cmidrule(lr){2-3} \cmidrule(lr){4-5}
& DSC & NSD & DSC & NSD \\
\cmidrule(lr){2-3} \cmidrule(lr){4-5}
Rand. U-Net 
& 0.022$\pm$0.005 & 0.033$\pm$0.005 
& 0.004$\pm$0.005 & 0.007$\pm$0.002 \\
All-classes U-Net* 
& \underline{0.399$\pm$0.101} & \textbf{0.311$\pm$0.143} 
& \textbf{0.292$\pm$0.217} & \textbf{0.216 $\pm$0.084} \\
\textsc{HyperUNet} \emph{(Ours)} 
& \textbf{0.500$\pm$0.119} & \underline{0.281$\pm$0.071}
& \underline{0.053$\pm$0.115} & \underline{0.053$\pm$0.108} \\
\emph{Conv. U-Net*} 
& \emph{0.859$\pm$0.034} & \emph{0.619$\pm$0.111}
& \emph{0.468$\pm$0.089} & \emph{0.364$\pm$0.056} \\
\end{tabular}%
}





\label{tab:results_mri_compact}
\end{table*}
For evaluation, we again use MSD tasks and report DSC and NSD. As shown in Fig.~\ref{tab:results_mri_compact}, the results mirror the CT experiments: \textsc{HyperUNet}s achieve strong performance on Heart and Brain Edema, demonstrating that the hypernetwork can generate meaningful parameters for targets unseen during training and even from an unseen imaging domain. Notably, they outperform the U-Net trained on all tasks on the Heart task ($0.50$ vs. $0.40$). The overall pattern, however, remains consistent: for large, well-defined structures, \textsc{HyperUNet}s clearly exceed the random baseline, whereas for small or complex targets (e.g., hippocampus, prostate peripheral zone) they offer only marginal gains over random initialisation.

\paragraph{Limited Data.}
Further, we study how \textsc{HyperUNet}s behave when more than a single reference pair is available. 
Instead of treating the synthesised decoder as a final segmenter, we view it as a task-informed initialisation that can be refined once additional supervision becomes available. 
Given $N$ annotated samples, we form a minimal train--test split by holding out 20\% of the data (at least one sample) for evaluation. 
A \textsc{HyperUNet} is synthesised from one reference pair in the training subset and then fine-tuned on the remaining training samples using standard backpropagation with early stopping. 
We evaluate multiple MSD tasks for $N \in \{2,4,8,16\}$, corresponding to $(1/1)$, $(3/1)$, $(6/2)$, and $(13/3)$ train/test splits.

As shown in Fig.~3, \textsc{HyperUNet}s can be effectively fine-tuned with as little as one additional annotated sample, where conventional U-Nets often overfit. 
Performance improves consistently with more data and gradually approaches fully supervised training.

Finally, we report MAML on the spleen task: despite differing shot counts (5 and 10), it performs comparably to \textsc{HyperUNet}. 
Appendix~\ref{appendix} (Tab.~\ref{tab:comp}) further compares \textsc{HyperUNet}s synthesised from five references to a 5-shot MAML baseline.

\paragraph{SOTA and similar approaches.}

\begin{table}[h]
\caption{\textbf{DSC [\%] of our \textsc{HyperUNet} and SOTA}~\cite{gao2025show} on MSD Pancreas.}
\centering
\begin{tabular}{ccccccc}
\textit{Ours} & \multicolumn{2}{c}{\emph{SAM-Style/Pos. Prompts}} & \multicolumn{4}{c}{\emph{In-context}}\\
\cmidrule(lr){2-3}\cmidrule(lr){4-7}
\textsc{HyperUNet} & SAM-Med2D & SAM-Med3D & SegGPT & UniverSeg & Tyche-IS & Iris\\
\cmidrule(lr){1-1}\cmidrule(lr){2-2}\cmidrule(lr){3-3}\cmidrule(lr){4-4}\cmidrule(lr){5-5}\cmidrule(lr){6-6}\cmidrule(lr){7-7}
 0.153 & 0.104 & 0.158 & 0.107 & 0.103 & 0.120 & 0.283\\
\end{tabular}
\label{tab:sota}
\end{table}

We compare \textsc{HyperUNet} to recent prompting-based and in-context learning approaches. 
In contrast to SAM-style prompting~\cite{cheng2023sam,wang2025sam} and ICL methods such as UniVerSeg~\cite{butoi2023universeg}, SegGPT~\cite{wang2023seggpt}, Tyche-IS~\cite{rakic2024tyche}, and Iris~\cite{gao2025show}, which perform reference-conditioned inference by matching each query to a fixed support set, \textsc{HyperUNet} synthesises a standalone, task-specific 3D U-Net. 
The generated model can be deployed directly or fine-tuned with standard supervised learning, without increasing inference-time memory or computation.

This benefit is reflected in Table~\ref{tab:sota}. On MSD Pancreas, a \textsc{HyperUNet} generated from a single reference pair reaches $0.153$ Dice, outperforming SAM-style and most ICL baselines. 
While Iris achieves higher performance ($0.283$), it relies on extensive support, query matching and retains the full support set at inference time. 
By compressing task information into model weights, \textsc{HyperUNet} enables efficient deployment and incremental refinement as more annotations become available. 
The efficiency of the proposed generate-once paradigm is further evidenced by the inference-time analysis reported in Table~\ref{tab:inf} (Appendix~\ref{appendix}). After an initial weight generation step ($1.77~s$), \textsc{HyperUNet} achieves a substantially lower per-sample inference time than prompting-based and in-context learning methods, reducing inference latency from $2.0~s$ (previous best) to $0.06~s$ per sample.

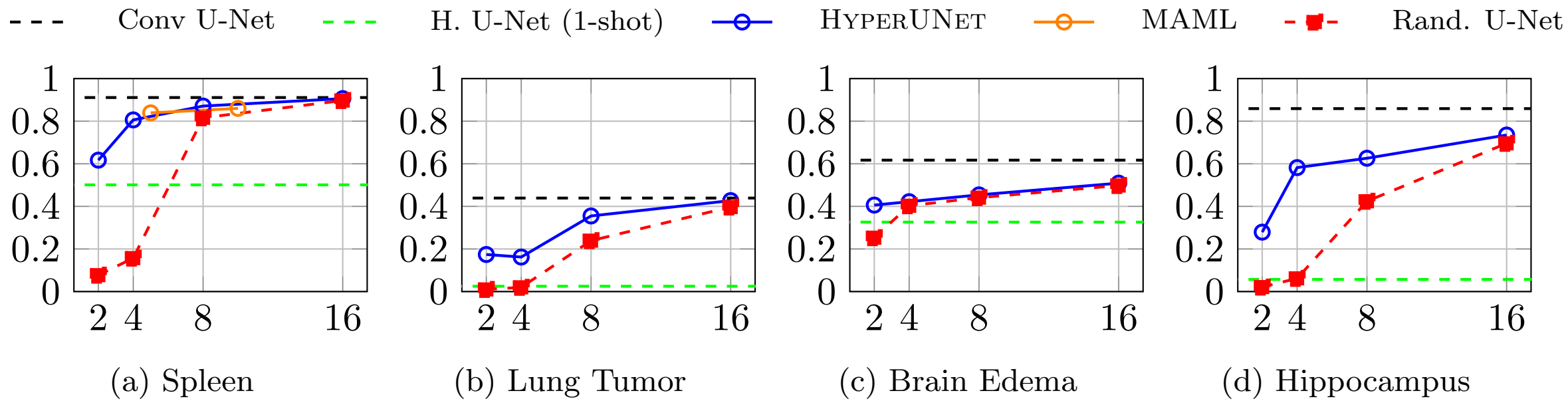
\begin{figure*}[t]
\centering

\begin{tikzpicture}
  \begin{axis}[
    hide axis,
    xmin=0, xmax=1,
    ymin=0, ymax=1,
    width=0pt,
    height=0pt,
    scale only axis,
    legend columns=5,     
    legend style={
      draw=none,
      fill=none,
      font=\scriptsize,
      column sep=1em,
      cells={anchor=west}
    },
    legend pos=north east,
    clip=false
  ]
    \addlegendimage{dashed, thick, black}
    \addlegendentry{Conv U-Net}

    \addlegendimage{color=green, dashed, thick}
    \addlegendentry{H. U-Net (1-shot)}

    \addlegendimage{color=blue, mark=o, thick, mark options={solid}}
    \addlegendentry{\textsc{HyperUNet}}

    \addlegendimage{color=orange, mark=o, thick, mark options={solid}}
    \addlegendentry{MAML}

    \addlegendimage{color=red, mark=square*, thick, dashed}
    \addlegendentry{Rand. U-Net}
  \end{axis}
\end{tikzpicture}


\begin{minipage}[t]{0.24\textwidth}
  \centering
  \begin{tikzpicture}
    \begin{axis}[
      width=4.5cm, height=3.7cm,
      ymin=0, ymax=1.0,
      xtick={2,4,8,16},
      grid=both
    ]
      \draw[dashed, thick, black] (axis cs:0,0.911) -- (axis cs:18,0.911);
      \draw[color=green, dashed, thick] (axis cs:0,0.501) -- (axis cs:18,0.501);
      \addplot[color=blue, mark=o, thick] coordinates {(2,0.617) (4,0.806) (8,0.871) (16,0.906)};
      \addplot[color=orange, mark=o, thick] coordinates {(5,0.839) (10,0.860)};
      \addplot[color=red, mark=square*, thick, dashed] coordinates {(2,0.077) (4,0.157) (8,0.816) (16,0.897)};
    \end{axis}
  \end{tikzpicture}

  {\footnotesize (a) Spleen}
\end{minipage}
\hfill
\begin{minipage}[t]{0.24\textwidth}
  \centering
  \begin{tikzpicture}
    \begin{axis}[
      width=4.5cm, height=3.7cm,
      ymin=0, ymax=1.0,
      xtick={2,4,8,16},
      grid=both
    ]
      \draw[dashed, thick, black] (axis cs:0,0.439) -- (axis cs:18,0.439);
      \draw[color=green, dashed, thick] (axis cs:0,0.025) -- (axis cs:18,0.025);

      \addplot[color=blue, mark=o, thick] coordinates {(2,0.174) (4,0.162) (8,0.355) (16,0.427)};
      \addplot[color=red, mark=square*, thick, dashed] coordinates {(2,0.01) (4,0.02) (8,0.239) (16,0.397)};
    \end{axis}
  \end{tikzpicture}

  {\footnotesize (b) Lung Tumor}
\end{minipage}
\hfill
\begin{minipage}[t]{0.24\textwidth}
  \centering
  \begin{tikzpicture}
    \begin{axis}[
      width=4.5cm, height=3.7cm,
      ymin=0, ymax=1.0,
      xtick={2,4,8,16},
      grid=both
    ]
      \draw[dashed, thick, black] (axis cs:0,0.617) -- (axis cs:18,0.617);
      \draw[color=green, dashed, thick] (axis cs:0,0.326) -- (axis cs:18,0.326);

      \addplot[color=blue, mark=o, thick] coordinates {(2,0.406) (4,0.422) (8,0.454) (16,0.509)};
      \addplot[color=red, mark=square*, thick, dashed] coordinates {(2,0.253) (4,0.403) (8,0.441) (16,0.498)};
    \end{axis}
  \end{tikzpicture}

  {\footnotesize (c) Brain Edema}
\end{minipage}
\hfill
\begin{minipage}[t]{0.24\textwidth}
  \centering
  \begin{tikzpicture}
    \begin{axis}[
      width=4.5cm, height=3.7cm,
      ymin=0, ymax=1.0,
      xtick={2,4,8,16},
      grid=both
    ]
      \draw[dashed, thick, black] (axis cs:0,0.859) -- (axis cs:18,0.859);
      \draw[color=green, dashed, thick] (axis cs:0,0.057) -- (axis cs:18,0.057);

      \addplot[color=blue, mark=o, thick] coordinates {(2,0.279) (4,0.582) (8,0.626) (16,0.735)};
      \addplot[color=red, mark=square*, thick, dashed] coordinates {(2,0.021) (4,0.062) (8,0.424) (16,0.696)};
    \end{axis}
  \end{tikzpicture}

  {\footnotesize (d) Hippocampus}
\end{minipage}

\caption{\textbf{DSC of \textsc{HyperUNet}s (blue) and Random U-Nets (red) trained using $N \in \{2, 4, 8, 16\}$ samples across several tasks.} A U-Net trained on the entire dataset (black) and a one-shot generated \textsc{HyperUNet} (green) are reported as baselines. For Spleen, MAML (orange) is reported as an additional baseline.}
\label{fig:results_limit}
\end{figure*}
\section{Conclusion}
In this work, we introduce \textsc{HyperUNet}, a hypernetwork framework that generates compact, task-specific 3D U-Nets from as little as a single annotated volume. By predicting decoder weights directly, rather than conditioning a shared backbone at the feature or logit level, \textsc{HyperUNet} produces fully instantiated segmentation models that can be deployed immediately or refined via lightweight fine-tuning, shifting task adaptation from gradient-based meta-learning to weight-level conditioning.
Experiments on CT and MRI tasks from the MSD show that one-shot weight generation is effective for anatomically homogeneous structures such as liver, spleen, and cardiac chambers, and that a hypernetwork trained solely on CT generalises to MRI by capturing modality-robust anatomical priors. For more complex or fine-grained structures, including hepatic vessels and tumours, performance degrades in the strict one-shot setting, reflecting an over-reliance on generic anatomical priors (e.g. location or coarse shape) that act as distractors for heterogeneous pathologies. However, treating the generated U-Nets as task-informed initialisations and fine-tuning on as few as 2-4 labelled volumes yields substantial gains, outperforming pretrained and randomly initialised baselines and closing much of the gap to fully supervised models.
Overall, \textsc{HyperUNet} offers an interesting alternative to gradient-based meta-learning by synthesising a complete task-specific segmentation model in a single forward pass, requiring only minimal additional optimisation. 
By explicitly separating shared priors from task-specific decoder synthesis, our method enables stable adaptation to previously unseen targets under severe annotation scarcity and occupies a distinct point in the design space. 
We view this work as an initial exploration of a complementary regime between in-context segmentation and fully trained task-specific models, and we hope it will motivate further research in this direction.
Future work will investigate task-dependent weighting of different priors (e.g., topology, location, texture, and intensity), richer context representations, alternative encoder architectures, and extensions to additional modalities such as PET and ultrasound.

\clearpage  

\section*{Acknowledgements} The authors received support by the ERC - project MIA-NORMAL 101083647, by DFG projects 513220538, 512819079, and by the state of Bavaria (HTA).
The authors gratefully acknowledge the scientific support and HPC resources provided by the Erlangen National High Performance Computing Center (NHR@FAU) of the Friedrich-Alexander-Universität Erlangen-Nürnberg (FAU) under the NHR project b143dc and b180dc. NHR funding is provided by federal and Bavarian state authorities. NHR@FAU hardware is partially funded by the German Research Foundation (DFG) - 440719683.

\bibliography{midl26_332}

\appendix
\section{Ablations}
\label{appendix}
To understand the contribution of individual components in our \textsc{HyperUNet} workflow, we conduct a series of ablation experiments. Each subsection isolates one specific design choice and reports its effect on segmentation performance across multiple tasks of the MSD dataset, where our \textsc{HyperUNet} previously achieved strong results. 

\subsection{Effect of Gradient Step on the Reference Example}
In our workflow, after generating a \textsc{HyperUNet} from the reference volume-mask pair $(x, y)$, we additionally perform a single gradient update with learning rate $\eta$ on this same reference pair. This step further optimises the generated model towards the reference example.\\
To assess the impact of this design choice, we compare the performance of \textsc{HyperUNet}s evaluated \emph{before} and \emph{after} applying the gradient step. 

\begin{table}[h!]
\centering
\caption[Ablation: Gradient step]{\textbf{Ablation with and without gradient step}. DSC $\uparrow$ and NSD $\uparrow$ across all CT and MRI tasks. Best configuration per task in \textbf{bold}.}
\setlength{\tabcolsep}{10pt}
\renewcommand{\arraystretch}{1.15}

\begin{tabular}{l l cc}
\toprule
\textbf{Task} & \textbf{Model} & \textbf{DSC $\uparrow$} & \textbf{NSD $\uparrow$} \\
\midrule

\multirow{2}{*}{Liver}
& w/o gradient step & 0.673$\pm$0.069 & 0.171$\pm$0.063 \\
& w/ gradient step \emph{(Ours)} & \textbf{0.772$\pm$0.063} & \textbf{0.490$\pm$0.080} \\
\midrule

\multirow{2}{*}{Spleen}
& w/o gradient step & 0.476$\pm$0.165 & 0.289$\pm$0.130 \\
& w/ gradient step \emph{(Ours)} & \textbf{0.501$\pm$0.169} & \textbf{0.316$\pm$0.115} \\
\midrule

\multirow{2}{*}{Hepatic Vessels}
& w/o gradient step & 0.281$\pm$0.102 & 0.306$\pm$0.100 \\
& w/ gradient step \emph{(Ours)} & \textbf{0.307$\pm$0.120} & \textbf{0.364$\pm$0.129} \\
\midrule

\multirow{2}{*}{Pancreas}
& w/o gradient step & \textbf{0.186$\pm$0.137} & \textbf{0.123$\pm$0.076} \\
& w/ gradient step \emph{(Ours)} & 0.153$\pm$0.138 & 0.102$\pm$0.078 \\
\midrule

\multirow{2}{*}{Brain Edema}
& w/o gradient step & 0.288$\pm$0.191 & 0.194$\pm$0.126 \\
& w/ gradient step \emph{(Ours)} & \textbf{0.326$\pm$0.208} & \textbf{0.217$\pm$0.150} \\
\midrule

\multirow{2}{*}{Heart}
& w/o gradient step & 0.457$\pm$0.053 & 0.254$\pm$0.035 \\
& w/ gradient step \emph{(Ours)} & \textbf{0.500$\pm$0.119} & \textbf{0.281$\pm$0.071} \\
\bottomrule

\end{tabular}
\label{tab:ablation_grad}
\end{table}

The DSC and NSD for both configurations are reported in Tab.~\ref{tab:ablation_grad}. 
For DSC, we observe a consistent increase in performance across most tasks, with the largest gain for the liver (0.772 vs. 0.673). 
In contrast, the pancreas task shows a slight decrease when applying the gradient step. 

For NSD, performance also improves for all tasks except the pancreas. 
Compared to the moderate DSC gains, NSD exhibits substantial improvements for the liver (0.490 vs. 0.171) and a notable increase for the hepatic vessels (0.364 vs. 0.306) when the gradient step is applied. 

\subsection{Without normalization}
The mask-averaged pooled encoder feature map $F_\mathrm{enc}$ serves as our global task encoding $z^{task}$, which adapts positional encodings $z^{pos}$ and generates the weights and bias for the final $1 \times 1 \times 1$ convolutional layer. \\
We compare two \textsc{HyperUNet} variants: one employing normalization\\
($z^{task} = \mathrm{LayerNorm}(\mathrm{MAP}(F_\mathrm{enc}))$) and one using the raw pooled features ($z^{task} = \mathrm{MAP}(F_\mathrm{enc})$).

\begin{table}[h!]
\centering
\caption[Ablation: Normalization]{\textbf{Ablation with vs.\ without normalization}. DSC $\uparrow$ and NSD $\uparrow$ across all CT and MRI tasks. Best configuration per task in \textbf{bold}.}
\setlength{\tabcolsep}{10pt}
\renewcommand{\arraystretch}{1.15}

\begin{tabular}{l l cc}
\toprule
\textbf{Task} & \textbf{Model} & \textbf{DSC $\uparrow$} & \textbf{NSD $\uparrow$} \\
\midrule

\multirow{2}{*}{Liver}
& w/o norm & \textbf{0.844$\pm$0.056} & 0.411$\pm$0.070 \\
& w/ norm \emph{(Ours)} & 0.772$\pm$0.063 & \textbf{0.490$\pm$0.080} \\
\midrule

\multirow{2}{*}{Spleen}
& w/o norm & 0.383$\pm$0.131 & 0.196$\pm$0.070 \\
& w/ norm \emph{(Ours)} & \textbf{0.501$\pm$0.169} & \textbf{0.316$\pm$0.115} \\
\midrule

\multirow{2}{*}{Hepatic Vessels}
& w/o norm & 0.014$\pm$0.041 & 0.017$\pm$0.048 \\
& w/ norm \emph{(Ours)} & \textbf{0.307$\pm$0.120} & \textbf{0.364$\pm$0.129} \\
\midrule

\multirow{2}{*}{Pancreas}
& w/o norm & 0.022$\pm$0.049 & 0.018$\pm$0.031 \\
& w/ norm \emph{(Ours)} & \textbf{0.153$\pm$0.138} & \textbf{0.102$\pm$0.078} \\
\midrule

\multirow{2}{*}{Brain Edema}
& w/o norm & 0.015$\pm$0.036 & 0.025$\pm$0.239 \\
& w/ norm \emph{(Ours)} & \textbf{0.326$\pm$0.208} & \textbf{0.217$\pm$0.150} \\
\midrule

\multirow{2}{*}{Heart}
& w/o norm & 0.053$\pm$0.130 & 0.137$\pm$0.025 \\
& w/ norm \emph{(Ours)} & \textbf{0.500$\pm$0.119} & \textbf{0.281$\pm$0.071} \\
\bottomrule

\end{tabular}
\label{tab:ablation_norm}
\end{table}

We report the DSC and NSD of both configurations in Tab.~\ref{tab:ablation_norm}. 
For liver and spleen, both relatively large and homogeneous organs, DSC and NSD remain of roughly similar magnitude across both configurations. 
For the remaining tasks, however, we observe a substantial degeneration of performance when no normalisation is applied. 
On the hepatic vessels, pancreas, brain oedema, and heart tasks, both DSC and NSD drop to values close to zero, indicating diffuse predictions that fail to capture underlying anatomical patterns.

\subsection{Task-dependent positional encodings}
We hypothesise that enriching positional encodings with task-specific information enables the \textsc{HyperUNet} to assign task-dependent roles and importance to different convolutional filters in the final network.\\ 
Concretely, we define 
\[
z^* = \mathrm{MLP}\big(\mathrm{Concat}[z^{task}, z^{pos}]\big),
\]
where $z^{task}$ is the global task encoding and $z^{pos}$ the positional encoding. 
To test this hypothesis, we compare two \textsc{HyperUNet} variants: 
(i) one employing task-dependent positional encodings $z^*$ as defined above, and 
(ii) one using purely positional encodings $z^* = z^{pos}$ without any task information. 
\begin{table}[h!]
\centering
\caption[Ablation: Task-specific positional encodings]{\textbf{Ablation with vs.\ without task-specific positional encodings}. DSC $\uparrow$ and NSD $\uparrow$ across all CT and MRI tasks. Best results per task in \textbf{bold}.}
\setlength{\tabcolsep}{10pt}
\renewcommand{\arraystretch}{1.15}

\begin{tabular}{l l cc}
\toprule
\textbf{Task} & \textbf{Model} & \textbf{DSC $\uparrow$} & \textbf{NSD $\uparrow$} \\
\midrule

\multirow{2}{*}{Liver}
& Regular PE & 0.654$\pm$0.151 & 0.158$\pm$0.055 \\
& Task-specific PE \emph{(Ours)} & \textbf{0.772$\pm$0.063} & \textbf{0.490$\pm$0.080} \\
\midrule

\multirow{2}{*}{Spleen}
& Regular PE & 0.376$\pm$0.165 & 0.198$\pm$0.081 \\
& Task-specific PE \emph{(Ours)} & \textbf{0.501$\pm$0.169} & \textbf{0.316$\pm$0.115} \\
\midrule

\multirow{2}{*}{Hepatic Vessels}
& Regular PE & 0.130$\pm$0.130 & 0.170$\pm$0.156 \\
& Task-specific PE \emph{(Ours)} & \textbf{0.307$\pm$0.120} & \textbf{0.364$\pm$0.129} \\
\midrule

\multirow{2}{*}{Pancreas}
& Regular PE & 0.116$\pm$0.118 & 0.082$\pm$0.069 \\
& Task-specific PE \emph{(Ours)} & \textbf{0.153$\pm$0.138} & \textbf{0.102$\pm$0.078} \\
\midrule

\multirow{2}{*}{Brain Edema}
& Regular PE & 0.295$\pm$0.197 & \textbf{0.246$\pm$0.132} \\
& Task-specific PE \emph{(Ours)} & \textbf{0.326$\pm$0.208} & 0.217$\pm$0.150 \\
\midrule

\multirow{2}{*}{Heart}
& Regular PE & 0.292$\pm$0.130 & 0.204$\pm$0.059 \\
& Task-specific PE \emph{(Ours)} & \textbf{0.500$\pm$0.119} & \textbf{0.281$\pm$0.071} \\
\bottomrule

\end{tabular}
\label{tab:ablation_pe}
\end{table}

The DSC and NSD for both configurations are reported in Tab.~\ref{tab:ablation_pe}.\\
Across all tasks, using task-specific positional embeddings gives a higher DSC than using task-agnostic ones. The performance gain is most notable for Heart (0.500 vs 0.292) and Hepatic Vessels (0.307 vs. 0.130), and smallest for Brain Edema (0.326 vs 0.295). 
For the NSD, we see a similar trend. The NSD is higher in almost all tasks, except the brain edema where the difference, however, is small (0.217 vs. 0.246). Similar to our previous ablation on the gradient step, we again see a signifant improvement of NSD for the Liver (0.490 vs. 0.158), Hepatic Vessels (0.364 vs 0.170) and Spleen (0.316 vs. 0.198) tasks.

\subsection{MAML vs \textsc{HyperUNet} for $5$ Reference pairs per Organ}

\begin{table}[H]
\centering
\begin{tabular}{lcccc}
\toprule
Model & Spleen & Liver & R Kidney & L Kidney \\
\midrule
MAML~\cite{alsaleh2024few}   & 0.839 & 0.903 & 0.775 & 0.870 \\
\textsc{HyperUNet} (\emph{Ours}) & 0.821 & 0.865 & 0.753 & 0.801 \\
\bottomrule
\end{tabular}
\caption{\textbf{DSC of our \textsc{HyperUNet} and an MAML approach using $5$ reference pairs} per organ.}
\label{tab:comp}
\end{table}

In Tab.~\ref{tab:comp}, MAML achieves slightly higher scores on a small set of closely related abdominal organs, a setting known to favour gradient-based meta-learning. \textsc{HyperUNet} is trained across more than 100 heterogeneous tasks spanning multiple organs and modalities, and does not rely on narrow task similarity, highlighting its broader applicability and complementary strengths.

\subsection{Sensitivity Analysis}
To assess \textsc{HyperUNet}'s sensitivity to the sampled reference pair, we conduct a small-scale study on a held-out test set of 20 samples from the MSD Spleen dataset. Specifically, we evaluate the test set using eight different \textsc{HyperUNet} instances, each synthesised from a distinct reference pair. The resulting DSC values range from 0.409 to 0.557, with a mean performance of $0.491 \pm 0.042$ across reference pairs. While this confirms an expected dependence on the chosen reference, the observed variability remains moderate and does not raise immediate concerns.

\subsection{Inference time}
Table \ref{tab:inf} reports the inference time of \textsc{HyperUNet} and related approaches.
\begin{table}[H]
    \centering
    \begin{tabular}{l r}
        \toprule
        \textbf{Method} & \textbf{Inference Time (s)} \\
        \midrule
        UniverSeg    & 659.4  \\
        SAM-Med2D      & 648.4  \\
        SAM-Med3D      & 15.2   \\
        Iris    & 2.0   \\
        \textsc{HyperUNet} \emph{(Ours)}    & 1.77 (Generation) / 0.06 (Inference)\\
        \bottomrule
    \end{tabular}
    \label{tab:inf}
    \caption{Empirical
measurements of inference time on one NVIDIA A100 GPU. The image size is processed to 128 × 128 × 128 for inference. For \textsc{HyperUNet} the time for U-Net generation and U-Net inference is reported.
}
\end{table}
\end{document}